\documentclass{article}
\PassOptionsToPackage{numbers,sort&compress}{natbib}
\usepackage[final]{tackling_climate_workshop_style}
\usepackage[utf8]{inputenc}
\usepackage[T1]{fontenc}
\usepackage{hyperref}
\usepackage{url}
\usepackage{booktabs}
\usepackage{amsfonts}
\usepackage{microtype}
\usepackage{enumitem}
\usepackage{amsmath}
\usepackage{graphicx}
\graphicspath{{figures/}}

\newcommand{\WEAVE}{WEAVE}
\newcommand{\WEAVEfull}{\underline{{W}}eather-field \underline{{E}}xchange for \underline{{A}}tmospheric \underline{{V}}ariable \underline{{E}}stimation}

\title{WEAVE: Weather-field Exchange for Atmospheric Variable Estimation}
\author{%
Anurup Naskar$^{1}$ \quad Pritthijit Nath$^{2}$ \vspace{0.2cm} \\
$^{1}$Courant Institute of Mathematical Sciences (CIMS), New York University \\
$^{2}$Department of Applied Mathematics and Theoretical Physics, University of Cambridge\\
\texttt{an4462@nyu.edu}; \texttt{pn341@cam.ac.uk}
}

\hypersetup{hidelinks,pdftitle={WEAVE: Weather-field Exchange for Atmospheric Variable Estimation},pdfauthor={Anurup Naskar and Pritthijit Nath}}

\begin{document}
\maketitle

\enlargethispage{1\baselineskip}

\vspace{-5mm}

\begin{abstract}
Temperature, water-vapour mixing ratio, and vertical velocity all describe the same evolving atmosphere, yet are commonly denoised independently. In this study, we introduce \WEAVE{} (\WEAVEfull{}), a residual DnCNN that combines field-specific pathways with shared multivariate context to reconstruct all three fields simultaneously. On processed ARM Southern Great Plains observations under matched 40\% Gaussian corruption, \WEAVE{} outperforms a training-calibrated anisotropic Gaussian smoother across every displayed field and metric. Compared with three independent DnCNN baselines, the simultaneous model improves mixing-ratio RMSE, PSNR, and SSIM across all five matched initialisations, while temperature improves in three of five initialisations for RMSE and PSNR and in all five for SSIM. \WEAVE{} also uses 11.23\% fewer parameters, achieves stronger spectral fidelity across all three fields, and retains information relevant to a vertical-motion regime diagnostic. Together, these results show that shared atmospheric structure can improve denoising with lower model complexity, providing a promising foundation for future multivariate reconstructions of atmospheric observations.
\end{abstract}

\section{Introduction}

Ground-based remote sensing resolves the lower atmosphere more continuously than sparse in situ sampling. Temperature ($T$) constrains stratification, water-vapour mixing ratio ($q$) describes moisture structure, and Doppler-lidar vertical velocity ($w$) captures coherent upward and downward motion~\cite{mather_arm_2013,sisterson_sgp_2016,newsom_profiles_2018,lothon_velocity_2009}. However, these measurements are affected by instrument noise, retrieval artefacts, and other sources of contamination. The coupled time--height evolution of these variables supports numerous studies of boundary-layer transport and convection, making it important to suppress such noise without erasing the variability that gives the observations their scientific value.

Conventional smoothing raises signal-to-noise ratio by sacrificing resolution, whereas ML-based denoising methods can adapt to local structure~\cite{selmer_lidar_2024}. We therefore include a training-calibrated anisotropic Gaussian smoother as a classical reference, alongside other denoising methods. Most ML approaches process each atmospheric variable independently, despite temperature, moisture, and vertical motion being simultaneous expressions of a coupled atmospheric state. Simply sharing all features is not necessarily safe either, because the fields have different physical ranges, intermittency, and spectral structure.

To exploit shared atmospheric information while preserving field-specific structure, we introduce \WEAVE{}, which combines field-specific encoders and residual heads with a shared atmospheric-context trunk. Field-specific pathways preserve each variable's identity, while the shared trunk interlaces complementary information from temperature, moisture, and vertical motion before reconstruction. The proposed model reconstructs $T$, $q$, and $w$ simultaneously in a single forward pass, allowing the three fields to be evaluated within a common multivariate framework. We compare it with three independent denoising CNNs (DnCNNs) across five matched initialisations and examine both pointwise reconstruction and coordinate-stitched spectra. A compact vertical-motion analysis then assesses whether the reconstructed fields remain useful for a process-relevant regime diagnostic.

\enlargethispage{1\baselineskip}

\section{Data and methods}

\textbf{Atmospheric observations:}~%
We use co-located products from the Atmospheric Radiation Measurement (ARM) Southern Great Plains (SGP) site during June 2020. The observations have a nominal cadence of 10-s and are represented as overlapping $64\times64$ time--height patches after quality control and deterministic gap completion. Training and model development use 1--20 June, while 564 patches from
21--22 June form a temporally held-out test set. The June 21--22 data are
excluded from gradient-based training and from model-selection decisions. We constrain each variable to a prescribed physical range: $T \in[220,330]$~K, $q \in[0,40]$~g\,kg$^{-1}$, and $w \in[-5,5]$~m\,s$^{-1}$.

\begin{figure}[!h]
\centering
\makebox[\textwidth][c]{\includegraphics[width=1.1\linewidth]{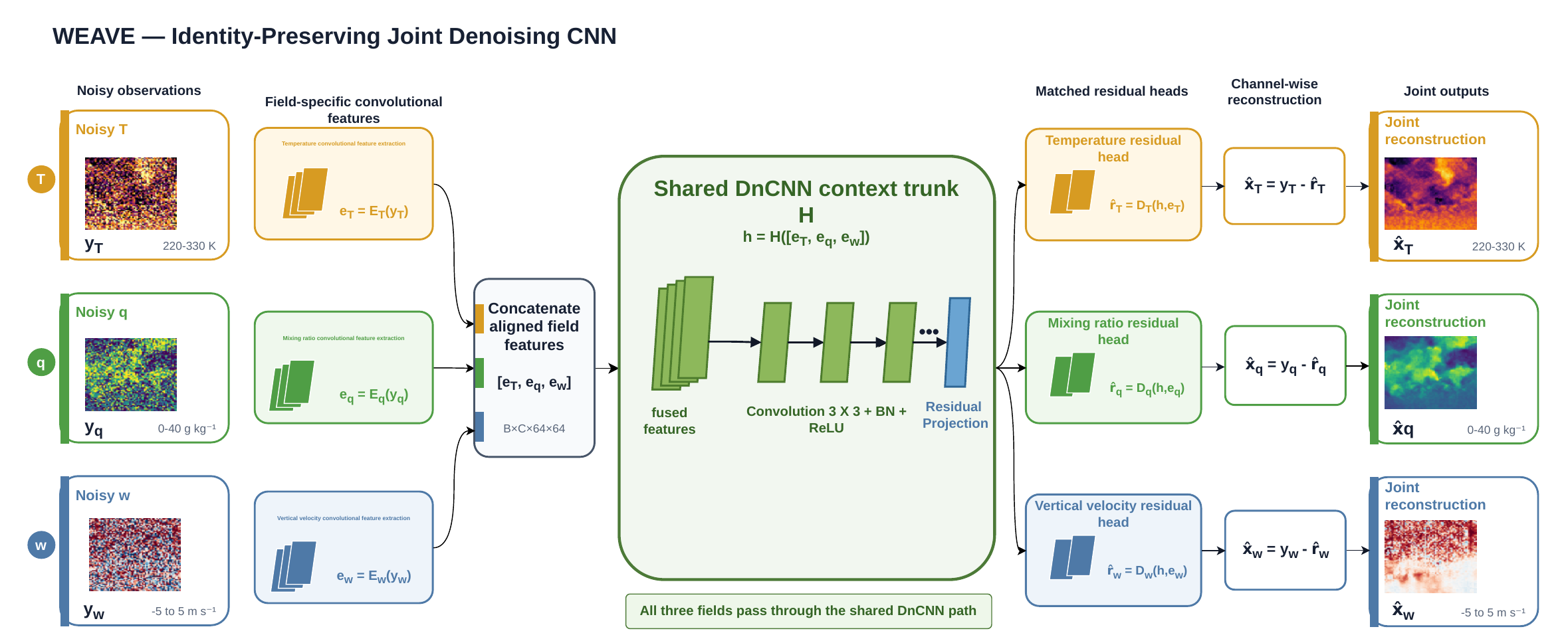}}
\caption{\WEAVE{} reconstructs temperature, mixing ratio, and vertical velocity simultaneously. Field-specific encoders and residual heads surround a shared atmospheric-context trunk, and each output subtracts only its matched predicted residual.}
\label{fig:architecture}
\end{figure}

\textbf{Corruption and residual reconstruction:}~%
Let $z_{p,k}$ denote the quality-controlled and gap-completed field $k$ in physical units over a physical space $p$. We define $\Pi_{[a_k,b_k]}$ as element-wise clipping to the prescribed range $[a_k,b_k]$, giving the processed reference $x_{p,k}=\Pi_{[a_k,b_k]}(z_{p,k})$. Controlled Gaussian corruption is then applied before a second range projection:
\begin{equation}
y_{p,k}
=
\Pi_{[a_k,b_k]}
\left(x_{p,k}+\epsilon_k\right),
\qquad
\epsilon_k
\sim
\mathcal N
\left(
0\;
[\alpha(b_k-a_k)]^2
\right),
\qquad
\alpha=0.4
\label{eq:corruption}
\end{equation}
where $\epsilon_k$ is sampled independently at each time--height location. The affine map $S_k(v)=(v-a_k)/(b_k-a_k)$ gives the normalised reference $x_k=S_k(x_{p,k})$, corrupted input $y_k=S_k(y_{p,k})$, and effective residual $r_k=y_k-x_k$. As clipping is part of the corruption process, $r_k$ need not equal the originally sampled perturbation.

To reconstruct this effective residual, each field-specific encoder forms $e_k=E_k(y_k)$, and the shared atmospheric-context trunk combines the three representations as $h=H([e_T,e_q,e_w])$. The corresponding residual head $D_k$ then predicts the residual for field $k$, which is subtracted from the corrupted input to obtain the reconstruction. Training jointly minimises the residual reconstruction error across all three fields:
\begin{equation}
\widehat r_k=D_k(h,e_k)
\qquad
\widehat x_k=y_k-\widehat r_k
\qquad
\mathcal L=
\frac{1}{3N\,64^2}
\sum_{n=1}^{N}
\sum_{k\in\{T,q,w\}}
\left\|
\widehat r_k^{(n)}
-
\left(y_k^{(n)}-x_k^{(n)}\right)
\right\|_2^2
\label{eq:weave}
\end{equation}
where $\widehat r_k$ is the predicted residual and $\widehat x_k$ is the resulting normalised reconstruction. Here, $N$ is the number of patches in the batch, $n$ indexes individual patches, and $\|\cdot\|_2^2$ sums the squared error over the $64\times64$ time--height grid. The encoders and skip connections retain field-specific information, while the shared trunk makes simultaneous observations from the other fields available as context (Figure~\ref{fig:architecture}). Physical reconstructions are obtained through $\widehat x_{p,k}=S_k^{-1}(\widehat x_k)$ without additional postprocessing. Under population squared-error risk, the optimum is the conditional mean $\widehat{\mathbf x}^{*}=\mathbb E[\mathbf x\mid\mathbf y]$ of the joint clean fields given the corrupted observations, although a finite network may not attain it~\cite{zhang_dncnn_2017}.

\textbf{Training and evaluation:}~%
\WEAVE{} and the Single models use identical training and test patches, corruption arrays, and matched initialisations. All networks use the predetermined epoch-500 checkpoint. The classical reference applies an anisotropic Gaussian kernel in physical coordinates, with field-specific height--time scales selected by minimising MSE on 1--20 June and frozen across all five evaluation streams. Reconstruction is evaluated using physical RMSE, PSNR, and SSIM~\cite{wang_ssim_2004}. Temporal variability is assessed after overlapping predictions are returned to their recorded coordinates using unsmoothed Hann periodograms~\cite{welch_psd_1967} at each retained range gate. \WEAVE{} contains 1,480,867 trainable parameters, 11.23\% fewer than the 1,668,291 parameters across the three independent DnCNNs. ~\ref{app:protocol} provides full protocol, implementation, and cross-field control details.

\section{Results}

\textbf{Reconstruction across the three fields:}~%
The classical comparison tests whether learnt denoising adds value beyond field-specific smoothing. \WEAVE{} outperforms Gaussian for every displayed mean in Table~\ref{tab:main-reconstruction} and for RMSE, PSNR, and SSIM across all five streams in each field. Spectral fidelity is also higher in every stream for $T$ and $q$, and in three of five for $w$. These results support learnt reconstruction under the evaluated corruption without claiming broader necessity for deep learning.

\begin{figure*}[!h]
\centering
\includegraphics[width=0.9\textwidth]{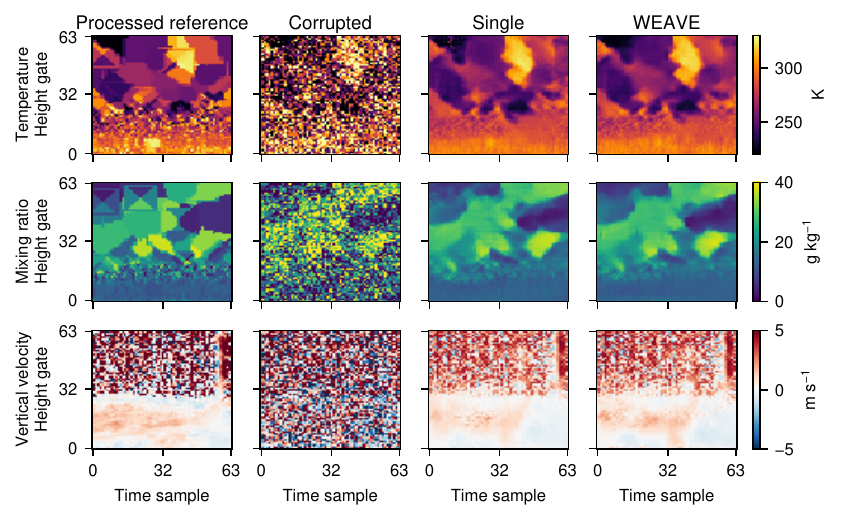}
\vspace{-6pt}
\caption{Representative 40\% corruption reconstructions for temperature, mixing ratio, and vertical velocity on fixed physical scales. The case was selected independently of reconstruction performance.}
\label{fig:reconstruction}
\vspace{-6pt}
\end{figure*}

Against the independent DnCNN baseline, \WEAVE{} achieves a favourable mean in 11 of 12 comparisons. For mixing ratio $q$, all three reconstruction metrics improve in every initialisation, with mean RMSE decreasing from $5.67$ to $4.64$~g\,kg$^{-1}$ and SSIM increasing from $0.344$ to $0.388$. Temperature RMSE and PSNR improve in three of five initialisations and SSIM in all five, while vertical-velocity performance remains comparable. Shared context therefore provides substantial moisture gains and clear temperature improvements while retaining competitive vertical-velocity reconstruction in a single network, as illustrated by the representative 40\% reconstruction in Figure~\ref{fig:reconstruction}.

\begin{table*}[!h]
\centering
\caption{Reconstruction at 40\% corruption, reported as mean $\pm$ sample standard deviation across five matched streams. Gaussian is a calibrated anisotropic smoother, Single comprises three independent DnCNNs, and \WEAVE{} reconstructs all fields jointly. Spectral fidelity is reported as a percentage. Bold marks the best mean within each field.} \vspace{2mm}
\label{tab:main-reconstruction}
\small
\setlength{\tabcolsep}{4.5pt}
\renewcommand{\arraystretch}{0.90}
\begin{tabular*}{\textwidth}{@{\extracolsep{\fill}}c l r r r r@{}}
\toprule
Field & Method & RMSE $\downarrow$ & PSNR $\uparrow$ & SSIM $\uparrow$ & Spectral fidelity $\uparrow$ \\
\midrule
$T$ & Gaussian & 17.93 $\pm$ 0.01 & 15.76 $\pm$ 0.00 & 0.253 $\pm$ 0.000 & 96.7 $\pm$ 0.1 \\
    & Single & 16.73 $\pm$ 2.08 & 16.41 $\pm$ 1.05 & 0.427 $\pm$ 0.010 & 97.4 $\pm$ 0.2 \\
    & \WEAVE{} & \textbf{15.28 $\pm$ 0.04} & \textbf{17.14 $\pm$ 0.02} & \textbf{0.434 $\pm$ 0.005} & \textbf{97.7 $\pm$ 0.1} \\
\midrule
$q$ & Gaussian & 6.13 $\pm$ 0.00 & 16.29 $\pm$ 0.01 & 0.211 $\pm$ 0.000 & 73.6 $\pm$ 0.2 \\
    & Single & 5.67 $\pm$ 1.08 & 17.09 $\pm$ 1.58 & 0.344 $\pm$ 0.009 & 82.6 $\pm$ 2.5 \\
    & \WEAVE{} & \textbf{4.64 $\pm$ 0.02} & \textbf{18.71 $\pm$ 0.04} & \textbf{0.388 $\pm$ 0.010} & \textbf{86.5 $\pm$ 0.4} \\
\midrule
$w$ & Gaussian & 2.41 $\pm$ 0.00 & 12.37 $\pm$ 0.00 & 0.257 $\pm$ 0.000 & 71.2 $\pm$ 0.2 \\
    & Single & 1.98 $\pm$ 0.01 & 14.06 $\pm$ 0.04 & \textbf{0.489 $\pm$ 0.003} & 71.4 $\pm$ 0.6 \\
    & \WEAVE{} & \textbf{1.98 $\pm$ 0.00} & \textbf{14.07 $\pm$ 0.01} & 0.488 $\pm$ 0.003 & \textbf{71.7 $\pm$ 0.6} \\
\bottomrule
\end{tabular*}
\end{table*}

\begin{figure*}[!h]
\centering
\includegraphics[width=0.9\textwidth]{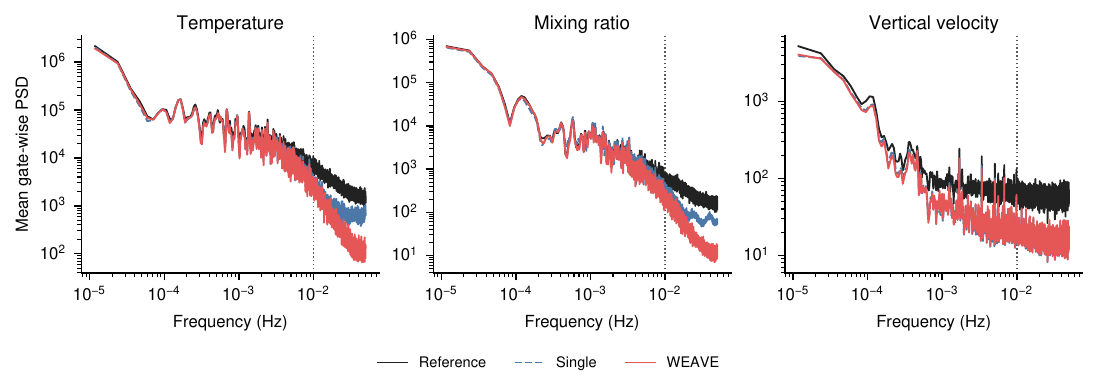}
\vspace{-6pt}
\caption{Coordinate-stitched spectra for the same representative 40\% case, averaged over the first 64 range gates using unsmoothed Hann periodograms. The vertical line marks 0.01~Hz.}
\label{fig:psd}
\vspace{-6pt}
\end{figure*}

\textbf{Temporal variability and regime preservation:}~%
Spectral fidelity favours \WEAVE{} over both references in mean across all three fields (Table~\ref{tab:main-reconstruction}), with the coordinate-stitched spectra shown in Figure~\ref{fig:psd}. 

The vertical-velocity reconstruction also preserves information relevant to a downstream regime diagnostic (shown in Table~\ref{tab:appendix-evaluation}). Using three fixed K-means centres fitted to the processed reference, \WEAVE{} improves macro-F1 from 0.6742 to 0.6753, balanced accuracy from 0.6455 to 0.6465, and adjusted Rand index from 0.4393 to 0.4402, with each improvement occurring in three of five initialisations.

\clearpage

\vspace*{-1cm}
\enlargethispage{\baselineskip}

\section{Climate relevance and conclusion}

As co-located temperature, moisture, and vertical-motion profiles characterise the coupled thermodynamic and dynamical evolution of the lower atmosphere, they support studies of boundary-layer development and shallow-to-deep convective transition~\cite{zhang_transition_2010}. \WEAVE{} reconstructs these joint time--height fields while retaining cross-variable relationships. Its fixed K-means diagnostic partitions vertical velocity into downward, near-neutral, and upward regimes, enabling temperature and moisture structure to be examined conditionally on vertical motion. These relationships support process-oriented analysis and can also be derived from climate-model output to assess observed thermodynamic--dynamical coupling.

Within the present ARM SGP experiments, \WEAVE{} outperforms classical smoothing across the displayed pointwise and spectral metrics, with its strongest gains over independent DnCNNs for moisture and temperature, while reconstructing all three fields with 11.23\% fewer parameters. Architecture controls support the field-specific/shared design, while auxiliary-field experiments indicate use of aligned cross-variable information. Reconstruction fidelity can be assessed through complementary pointwise, structural, spectral, and residual diagnostics before downstream use. Future work will extend evaluation across longer records, seasons, sites, and convectively active environments, including independently identified shallow-to-deep convective events. Together, these results show that shared atmospheric structure can improve denoising with lower model complexity, providing a promising foundation for multivariate reconstruction of atmospheric observations.

\begin{ack}
P. Nath was supported by the \href{https://ai4er-cdt.esc.cam.ac.uk/}{UKRI Centre for Doctoral Training in Application of Artificial Intelligence to the study of Environmental Risks} [EP/S022961/1]. 
\end{ack}

\section*{LLM usage disclosure}
The authors acknowledge the use of AI language models, specifically OpenAI Codex and ChatGPT~(GPT-5.6 Sol), during the preparation of this work. These tools were used to assist in software development, polish language usage and improve the overall clarity of the manuscript. All AI-generated content was reviewed, verified, and edited by the authors to ensure accuracy and appropriateness.

\bibliographystyle{vancouver}
\bibliography{bibliography}

\clearpage
\appendix
\renewcommand\thesection{Appendix \Alph{section}}
\renewcommand\thesubsection{\Alph{section}.\arabic{subsection}}
\renewcommand\thetable{\Alph{section}.\arabic{table}}
\renewcommand\thefigure{\Alph{section}.\arabic{figure}}
\renewcommand\theHtable{appendix.\Alph{section}.\arabic{table}}
\renewcommand\theHfigure{appendix.\Alph{section}.\arabic{figure}}
\setcounter{table}{0}
\setcounter{figure}{0}

\section{Protocol and supporting analyses}
\label{app:protocol}

All methods share the same source products, patch coordinates, normalisation ranges, corruption arrays, and evaluation arrays. Training is restricted to 1--20 June and evaluation to the temporally separated 21--22 June period, with all reported networks taken from the predetermined epoch-500 checkpoints for initialisations 17, 42, 101, 202, and 303. Within this common protocol, preprocessing, corruption, training, and evaluation proceed as follows:

\begin{enumerate}[leftmargin=*,label=\textbf{\arabic*.}]

\item \textbf{Inputs:} ARM SGP Raman-lidar \texttt{temperature} and \texttt{mr\_merged} from \nolinkurl{sgprlproftemp2newsC1.c0} and \nolinkurl{sgprlprofmr2newsC1.c0} plus elevation-corrected Doppler-lidar \texttt{radial\_velocity} from \nolinkurl{sgpdlfptC1.b1}; Doppler samples pass paired zero-valued quality flags and are matched to Raman times within 2~s

\item \textbf{Grid and patches:} Raman fields are interpolated to 66 levels from 30 to 3930~m at 60~m spacing; deterministic $64\times64$ patches use 9--11~s continuous sampling with stride 64 in time and vertical starts at indices 0 and 2

\item \textbf{Gap completion:} Valid-neighbour means use windows of $3$ $5$ $7$ $9$ $11$ and $15$ cells; remaining gaps are filled by interpolation in time then height followed by nearest-neighbour filling and the training-period mean when required; evaluation data never determine fallback means

\item \textbf{Corruption:} Gaussian perturbations are scaled by each field's physical range and applied before a second clipping step; near a physical boundary the effective learning residual can therefore differ from the sampled perturbation

\item \textbf{Training:} Single uses three 17-layer residual DnCNNs whereas \WEAVE{} uses field encoders of width 44 with a shared trunk of width 88 and matched residual heads; both use residual MSE with batch size 32 Adam at $10^{-3}$ cosine annealing flips with probability 0.5 and bfloat16 mixed precision

\item \textbf{Evaluation:} RMSE retains field-specific physical units while PSNR uses the prescribed physical range and SSIM the processed reference; coordinate-stitched Hann periodograms define spectral fidelity as the percentage of bins at $0<f\leq0.01$~Hz with epsilon-log discrepancy no greater than 0.5 in magnitude

\end{enumerate}

Under the same evaluation framework, the anisotropic Gaussian reference uses reflected boundaries and field-specific $(\sigma_h,\sigma_t)$ values of $(3,1)$ for $T$, $(2,3)$ for $q$, and $(0.75,0.75)$ for $w$. These values minimise training MSE over a fixed candidate grid, with near-ties resolved towards less smoothing. The implementation uses Python 3.10.18, PyTorch 2.6.0 with CUDA 12.4, NumPy 2.2.6, SciPy 1.15.3, and xarray 2025.6.1 on one NVIDIA RTX A5000/A6000 GPU. Training requires 0.62--0.74 GPU hours per Single model and 1.43--1.66 GPU hours per final-width \WEAVE{} model, excluding preparation and evaluation. 

\subsection{Evaluation summaries}

Table~\ref{tab:appendix-evaluation} summarises the per-stream Gaussian results and downstream regime diagnostic. Variation in the Gaussian results reflects only the matched corruption streams. For the regime diagnostic, fixed K-means centres fitted to processed-reference vertical velocity define downward, near-neutral, and upward regimes, so the reported scores measure agreement with this reference-derived partition rather than independently verified convective states.

\WEAVE{} achieves a higher displayed mean for five of the six regime metrics, although each gain occurs in only three of five matched initialisations, while near-neutral IoU is unchanged at the reported precision. The simultaneous model therefore does not improve every metric for every field, and the event-selected May stress test does not establish temporal generalisation.

\begin{table}[htbp]
\centering
\small
\caption{Supporting evaluation at 40\% corruption across five matched initialisations. Values are mean $\pm$ sample standard deviation. Spectral fidelity is a percentage, and higher downstream scores are favourable.}
\label{tab:appendix-evaluation}
{(a) Training-calibrated Gaussian reference}\\[2pt]
\setlength{\tabcolsep}{3.5pt}
\begin{tabular}{crrrr}
\toprule
Field & RMSE & PSNR & SSIM & Spectral fidelity \\
\midrule
$T$ & 17.930 $\pm$ 0.010 & 15.756 $\pm$ 0.005 & 0.2532 $\pm$ 0.0004 & 96.68 $\pm$ 0.07 \\
$q$ & 6.128 $\pm$ 0.005 & 16.295 $\pm$ 0.007 & 0.2114 $\pm$ 0.0002 & 73.65 $\pm$ 0.24 \\
$w$ & 2.4058 $\pm$ 0.0004 & 12.375 $\pm$ 0.002 & 0.2571 $\pm$ 0.0004 & 71.15 $\pm$ 0.20 \\
\bottomrule
\end{tabular}

\vspace{2mm}
{(b) Reference-derived vertical-motion regimes}\\[2pt]
\begin{tabular}{lrrc}
\toprule
Metric & Single & \WEAVE{}\\
\midrule
Macro-F1 & 0.6742 $\pm$ 0.0039 & \textbf{0.6753 $\pm$ 0.0020} \\
Balanced accuracy & 0.6455 $\pm$ 0.0032 & \textbf{0.6465 $\pm$ 0.0017}\\
Adjusted Rand index & 0.4393 $\pm$ 0.0023 & \textbf{0.4402 $\pm$ 0.0020}\\
Downward IoU & 0.2292 $\pm$ 0.0081 & \textbf{0.2315 $\pm$ 0.0040}\\
Near-neutral IoU & \textbf{0.6941 $\pm$ 0.0006} & \textbf{0.6941 $\pm$ 0.0006}\\
Upward IoU & 0.7097 $\pm$ 0.0014 & \textbf{0.7103 $\pm$ 0.0016}\\
\bottomrule
\end{tabular}
\end{table}

\subsection{Architecture and conditioning controls}

The controls in Table~\ref{tab:merged-controls} were run after the final configuration was fixed and did not influence model selection. The three panels examine capacity-matched pathway design, training-time alignment, and final-width context size. The structured model performs best across all displayed metrics in the capacity-matched control, although this single-initialisation experiment uses a narrower predecessor of the final model. Aligned context provides the clearest benefit for mixing ratio, while temperature and vertical velocity show smaller or metric-dependent changes. Overall, the controls support field-dependent use of auxiliary information rather than a universal gain.

\begin{table}[!h]
\centering
\small
\caption{Selected architecture and conditioning controls at 40\% corruption and epoch 500. Panels use one predetermined initialisation. Panel (a) reports RMSE, whereas panels (b) and (c) report field-specific MSE. Bold marks the favourable displayed value within each target and panel.} \vspace{2mm}
\label{tab:merged-controls}
{(a) Capacity-matched pathway control}\\[2pt]
\setlength{\tabcolsep}{2.5pt}
\begin{tabular}{clrrr}
\toprule
Field & Architecture & RMSE $\downarrow$ & PSNR $\uparrow$ & SSIM $\uparrow$ \\
\midrule
$T$ & Matched shared & 15.307 & 17.193 & 0.4229 \\
    & Structured & \textbf{15.168} & \textbf{17.274} & \textbf{0.4359} \\
$q$ & Matched shared & 4.575 & 19.044 & 0.3793 \\
    & Structured & \textbf{4.544} & \textbf{19.101} & \textbf{0.3914} \\
$w$ & Matched shared & 1.982 & 14.073 & 0.4825 \\
    & Structured & \textbf{1.974} & \textbf{14.107} & \textbf{0.4907} \\
\bottomrule
\end{tabular}

\vspace{2mm}
\begin{minipage}[t]{0.49\linewidth}
\centering
\small
{(b) Training-time alignment in the earlier shared model}\\[2pt]
\setlength{\tabcolsep}{1.5pt}
\begin{tabular}{clrrr}
\toprule
Target & Training inputs & MSE $\downarrow$ & PSNR $\uparrow$ & SSIM $\uparrow$ \\
\midrule
$T$ & $T,q,w$ aligned & 255.67 & 16.858 & 0.4070 \\
    & $T,q,w$ shuffled & 281.98 & 16.621 & 0.3735 \\
    & $T,q$ aligned & \textbf{243.13} & \textbf{17.126} & \textbf{0.4164} \\
$q$ & $T,q,w$ aligned & 23.96 & 18.612 & 0.3388 \\
    & $T,q,w$ shuffled & 34.70 & 17.927 & 0.3084 \\
    & $T,q$ aligned & \textbf{22.87} & \textbf{18.879} & \textbf{0.3648} \\
$w$ & $T,q,w$ aligned & 4.17 & 13.834 & 0.4382 \\
    & $T,q,w$ shuffled & 4.14 & 13.872 & 0.4681 \\
    & $T,w$ aligned & \textbf{4.03} & \textbf{13.985} & \textbf{0.4767} \\
\bottomrule
\end{tabular}
\end{minipage}
\hfill
\begin{minipage}[t]{0.49\linewidth}
\centering
\small
{(c) Context size in the final-width model}\\[2pt]
\setlength{\tabcolsep}{1.5pt}
\begin{tabular}{clrrr}
\toprule
Target & Available inputs & MSE $\downarrow$ & PSNR $\uparrow$ & SSIM $\uparrow$ \\
\midrule
$T$ & $T$ & \textbf{232.61} & \textbf{17.162} & 0.4307 \\
    & $T,q,w$ & 235.40 & 17.110 & \textbf{0.4320} \\
$q$ & $q$ & 22.66 & 18.488 & 0.3411 \\
    & $T,q,w$ & \textbf{21.70} & \textbf{18.677} & \textbf{0.3839} \\
$w$ & $w$ & \textbf{3.90} & \textbf{14.084} & \textbf{0.4916} \\
    & $T,q,w$ & 3.91 & 14.076 & \textbf{0.4916} \\
\bottomrule
\end{tabular}
\end{minipage}
\end{table}

\subsection{Reliance on aligned auxiliary fields}

Table~\ref{tab:context-reliance} perturbs the five frozen \WEAVE{} initialisations by deterministically reassigning complete auxiliary $64\times64$ patches, while keeping the target, reference, corruption, and checkpoint fixed. Positive differences therefore indicate an advantage for aligned inputs. The strongest reliance is on aligned moisture for temperature and aligned temperature for mixing ratio, with a smaller effect for vertical velocity. Because shuffled inputs may fall outside the training distribution, this analysis diagnoses fitted-model reliance rather than providing causal evidence for the training-time gain.

\begin{table}[htbp]
\centering
\small
\caption{Inference reliance on aligned auxiliary fields across five initialisations. Every difference favours aligned inputs in all five matched comparisons. MSE differences retain field-specific squared physical units.}
\label{tab:context-reliance}
\setlength{\tabcolsep}{1.5pt}
\begin{tabular}{clrrr}
\toprule
Target & Shuffled & $\Delta$MSE & $\Delta$PSNR & $\Delta$SSIM \\
\midrule
$T$ & $q$ & 8.788 & 0.160 & 0.0281 \\
    & $w$ & 0.836 & 0.016 & 0.0027 \\
    & $q,w$ & 9.732 & 0.177 & 0.0299 \\
$q$ & $T$ & 2.243 & 0.430 & 0.0672 \\
    & $w$ & 0.154 & 0.031 & 0.0027 \\
    & $T,w$ & 2.280 & 0.437 & 0.0684 \\
$w$ & $T$ & 0.009 & 0.010 & 0.0011 \\
    & $q$ & 0.008 & 0.009 & 0.0011 \\
    & $T,q$ & 0.016 & 0.018 & 0.0027 \\
\bottomrule
\end{tabular}
\end{table}

\end{document}